\documentclass[letterpaper]{article} % DO NOT CHANGE THIS
\usepackage{aaai23}  % DO NOT CHANGE THIS
\usepackage{times}  % DO NOT CHANGE THIS
\usepackage{helvet}  % DO NOT CHANGE THIS
\usepackage{courier}  % DO NOT CHANGE THIS
\usepackage[hyphens]{url}  % DO NOT CHANGE THIS
\usepackage{graphicx} % DO NOT CHANGE THIS
\usepackage{natbib}  % DO NOT CHANGE THIS AND DO NOT ADD ANY OPTIONS TO IT
\usepackage{caption} % DO NOT CHANGE THIS AND DO NOT ADD ANY OPTIONS TO IT
\usepackage{algorithm}
\usepackage{algorithmic}

\usepackage{newfloat}
\usepackage{listings}
\DeclareCaptionStyle{ruled}{labelfont=normalfont,labelsep=colon,strut=off} % DO NOT CHANGE THIS
\floatstyle{ruled}
\newfloat{listing}{tb}{lst}{}
\floatname{listing}{Listing}
\usepackage{svg}
\usepackage{tikz}
\usepackage{tikz-cd}
\usetikzlibrary{matrix, positioning}

\usepackage{amssymb}
\usepackage{amsmath}
\usepackage{amsthm}
\newtheorem{definition}{Definition}
\newtheorem{proposition}{Proposition}
\newtheorem{lemma}{Lemma}

\usepackage{multirow}
\usepackage{subcaption}

\newcommand{\bra}[1]{\langle\, #1 \, |}
\newcommand{\ket}[1]{|\, #1 \, \rangle}
\DeclareMathOperator{\atantwo}{atan2}
 
\newcommand{\cloud}[1]{\langle #1 \rangle}
\newcommand{\Euc}{\mathrm{Euc}}
\newcommand{\M}{\mathcal{M}}
\newcommand{\R}{\mathbb{R}}

\newcommand{\Z}{\mathbb{Z}}
\newcommand{\ph}{\varphi}

\newcommand{\src}[1]{{\mathrm{src}(#1)}}
\newcommand{\tgt}[1]{{\mathrm{tgt}(#1)}}
\newcommand{\silu}[1]{{\mathrm{silu}(#1)}}
\newcommand{\module}[1]{{[ #1]}}
\newcommand{\cell}{{\bar\gamma}}

\usepackage{mathtools, stmaryrd}
\usepackage{xparse} \DeclarePairedDelimiterX{\Iintv}[1]{\llbracket}{\rrbracket}{\iintvargs{#1}}
\NewDocumentCommand{\iintvargs}{>{\SplitArgument{1}{,}}m}
{\iintvargsaux#1} %
\NewDocumentCommand{\iintvargsaux}{mm} {#1\mkern1.5mu,\mkern1.5mu#2}

\usepackage{todonotes}

\title{Equivariant Message Passing Neural Network for Crystal Material Discovery}
\author{
    Astrid Klipfel\textsuperscript{\rm 1,2,3}\thanks{corresponding author},
    Olivier Peltre\textsuperscript{\rm 1,3},
    Najwa Harrati\textsuperscript{\rm 2},
    Yaël Fregier\textsuperscript{\rm 3},\\
    Adlane Sayede\textsuperscript{\rm 2},
    Zied Bouraoui\textsuperscript{\rm 1}
}
\affiliations{
   \textsuperscript{\rm 1} Univ. Artois, UMR 8188, Centre de Recherche en Informatique de Lens (CRIL), F-62300 Lens, France. \\
     \textsuperscript{\rm 2} Univ. Artois, UMR 8181, Unité de Catalyse et de Chimie du Solide (UCCS), F-62300 Lens, France.\\
     \textsuperscript{\rm 3} Univ. Artois, UR 2462, Laboratoire de Mathématiques de Lens (LML), F-62300 Lens, France.
}
\usepackage{bibentry}
\begin{document}

\maketitle

\begin{abstract}
Automatic material discovery with desired properties is a fundamental challenge for material sciences. Considerable attention has recently been devoted to generating stable crystal structures. While existing work has shown impressive success on supervised tasks such as property prediction, the progress on unsupervised tasks such
as material generation is still hampered by the limited extent to which the equivalent geometric representations of the same crystal are considered. To address this challenge, we propose EMPNN a periodic equivariant message-passing neural network that learns crystal lattice deformation in an unsupervised fashion. Our model equivalently acts on lattice according to the deformation action that must be performed, making it suitable for crystal generation, relaxation and optimisation. We present experimental evaluations that demonstrate the effectiveness of our approach. 
\end{abstract}

\section{Introduction}
Discovering thermodynamic stable materials with desired properties is a fundamental challenge for material sciences. Considerable attention has recently been devoted to crystalline (crystal) material generation. Crystals are involved everywhere in our modern society from metal alloys to semiconductors. Contrarily to organic molecules which are mostly composed of wide carbon chains with a limited variety of atoms, crystals are three-dimensional periodic structures composed of a wider variety of chemical bonds and atoms. The periodic structure is often represented as a parallelepiped tiling, a.k.a crystal lattice or unit cell.
  
Within the broad aim of automated stable (crystal) material discovery, various strategies mainly based on simulation or Machine Learning (ML) can be explored. Simulation allows the properties of a given structure to be predicted by applying physics laws while ML consists of modelling and predicting the physical properties. Notice that simulation can also be used for material relaxation, i.e.\ modifying a  structure to improve its stability. The success of ML has led to a paradigm shift in materials science. In particular, ML techniques are used for performing molecule design, modelling physical properties or at the early stage of material discovery. Recently, several works have been introduced to manipulate crystal structures, e.g. \cite{REN2022314,Long2021}. Most notably, models based on geometrically equivariant ML techniques such as Message Passing Neural Networks (MPNNs) have shown good performance in theoretical chemistry, in particular, on supervised tasks such as property predictions on both organic and crystalline structures, e.g. \cite{xie2021crystal,klicpera2022gemnet}. However, the majority of existing models are not fully equivariant, making them unsuitable for unsupervised tasks such as generation or representation learning. For example, the method from \cite{klicpera2022gemnet} is only equivariant to SO(3) (rotation group), making it not suitable for crystal lattice deformation where the shape of the structure is unknown in advance. To this end, some methods have been proposed to approximate Density Functional Theory (DFT) simulation using MPNNs for unsupervised tasks, e.g.\ \cite{PhysRevMaterials.6.033801,https://doi.org/10.48550/arxiv.2202.13947}. They rely on self-simulations to gather information about the interaction forces of a few specific structures to perform generation. However, discovering new materials requires a consequent amount of data to obtain out-of-distribution generalization, i.e.\ knowledge needed to generalise to unknown structures and perform arbitrary lattice deformation.

%To learn such knowledge one can not build larger databases of stable structure or employ data augmentation strategies. as only a small and restricted amount of chemical elements for which physics laws are the same everywhere are available. Notice that we can not rely on randomly generated structures as they lead in general to unstable structures.

We propose EMPNN an equivariant MPNN that acts on crystal lattice without any label from the interaction forces and stress tensors. Previous works already showed the advantage of using MPNN acting on atomic position for both organic molecules and crystals. But acting on crystal lattices without explicit stress tensors remains a challenging problem. Our model enforces a structuring bias adapted to crystals using group actions incorporated by the equivariance property of MPNN layers. To illustrate intuition, given a pair of atoms, if we know their interaction force in a given state, we can generalize this interaction to any other orientation as long as the state and the relative distance remain the same. Hence, we can take advantage of this property, and the equivariant representation to enhance the generalisation capability. This allows our model to equivalently act on crystal lattice according to the deformation action that needs to be performed. We consider equivariance with respect to the Euclidean group $Euc(3)$ and $\text{SL}_3(\Z)$ group. To the best of our knowledge, our model is the first general framework that formulates an equivariant MPNN on the periodic structures.  To demonstrate the effectiveness of our model, we propose a number of evaluation tasks to compare multiple equivariant MPNNs and losses. 
% In summary, our contributions are: 
% \begin{itemize}
%     \item As crystals are infinite collections of atoms, we need to work with finite descriptions. However, given a finite piece of a structure, our model should produce results that aren't specific and dependent on the piece of structure, but on the material itself. We introduced $SL(Z)$ to keep track of which pieces of structures are equivalent. 
%     \item Performing lattice deformation allows learning structural bias. Analogically to multi-label image classification where convolutions are equivariant by translation since the location of objects in an image doesn't change their labels, we introduced equivariant actions that don't change the properties by translation, rotation and repaving $SL(Z)$.
% \end{itemize}
 
%%%%%%%%%%%%%%%%%%%%%%%%%%%%%%%%%%%%%%%%%%%%%%%%%%%%%%%%%%%%%%%%%%%%%%%%%%%%%%%%%%%%%%%%%%%%%%%%%%%%%%%%%%%%%%%%%%%%%%%%%%%%%%%%%%%%%%%%%%%%%%%%%%%%%%%%%%%%%%%%%%%%%%%%%%

\section{Related works}

Within the area of automatic stable material discovery,
We can identify three classes of related work according to the molecular descriptors used to represent data.

\noindent\textbf{Fingerprint.} 
This class of methods uses handcrafted features of the materials. They are based on fingerprint representation that includes atomic positions and lattice parameters \cite{REN2022314}.  Additional information such as electronegativity, atomic radius or interatomic distances can also be incorporated, e.g. \  \cite{doi:10.1021/acscentsci.0c00426,DBLP:journals/corr/abs-1810-11203}. Those works mainly rely on Feedforward Neural Network (FFN) architectures to build Variational Autoencoder \cite{https://doi.org/10.48550/arxiv.1312.6114} or Generative Adversarial Networks \cite{NIPS2014_5ca3e9b1} to achieve generation or optimization tasks. However, 
fingerprints do not satisfy the uniqueness property, i.e.\ the same crystal can have different representations.
As FFNs are not equivariant to permutation, alternative representations of the same material can be processed differently. The same observation can be made for other group actions. Finally, existing models don't take into account periodicity.

\noindent\textbf{Voxel.} Offering a convenient way to represent data in 3-dimensional space, voxels allow encoding lattice parameters and atomic positions 
%(generally represented by their Cartesian coordinates) 
\cite{doi:10.1021/acs.jcim.0c00464,doi:10.1126/sciadv.aax9324,NOH20191370,Long2021}. However, voxel-based representation is limited since input data are by nature sparse and discontinuous in the space. Moreover, voxels do not take into account periodicity, which can lead to an edge effect. Finally, the aforementioned methods are not equivariant. As shown in section \ref{materials_equivariance}, there are multiple equivalent representations of a given material. Therefore, a set of equivalent representations may lead to inconsistent results. This is a clear limitation of voxel-based representation models.

\noindent\textbf{Graph-based Representation.}  Graph representation of materials can represent the local environment of each atom and structure periodicity. Recent works suggested using Graph Neural Networks (GNN) for materials \cite{xie2021crystal}. MPNNs allow to process sparse data and can be designed to be invariant or equivariant to many group actions. Most of the existing works are equivariant to $\text{SO}(3)$ \cite{klicpera2022gemnet} thanks to a spherical basis that allows us to predict lattice properties and perform simulations. However, these methods are not able to deform crystal lattices where the shape of the lattice is unknown in advance. In addition, these works are equivariant to subgroups of the Euclidian group but do not consider other group actions such as $\text{SL}_3(\mathbb{Z})$. Several methods have been proposed to approximate DFT simulation with GNN. These methods work by learning interaction forces and stress tensors to lower the total energy of a structure with methods analogue to DFT calculation \cite{Pickard_2011,PhysRevMaterials.6.033801,https://doi.org/10.48550/arxiv.2202.13947,https://doi.org/10.48550/arxiv.2012.02920}. These equivariant methods require a lot of additional information about interaction forces, which are not always available. They mainly use self-simulations to gather data, but only for a few specific structures. To discover new materials, we need a lot of data and cannot rely on randomly generated structures, as they lead in general to unstable structures.

%%%%%%%%%%%%%%%%%%%%%%%%%%%%%%%%%%%%%%%%%%%%%%%%%%%%%%%%%%%%%%%%%%%%%%%%%%%%%%%%%%%%%%%%%%%%%%%%%%%%%%%%%%%%%%%%%%%%%%%%%%%%%%%%%%%%%%%%%%%%%%%%%%%%%%%%%%%%%%%%
\begin{figure}[t]
    \begin{minipage}[c]{0.50\linewidth}
        \centering
        \includegraphics[width=\textwidth]{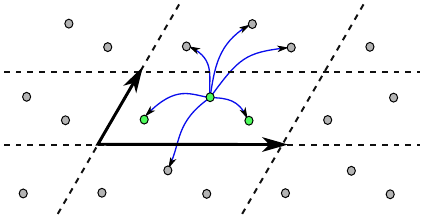}
    \end{minipage}
    \hfill
    \begin{minipage}[c]{0.30\linewidth}
        \centering
        \includegraphics[width=\textwidth]{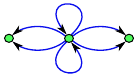}
    \end{minipage}
    \caption{Periodic structure represented as a lattice (in dotted lines). The multi-graph associated with a material (blue arrow) can overlap on the adjacent repetition of the lattice and a pair of nodes can have multiple connections.}
    \label{fig:material_graph}
\end{figure}
\section{Problem Setting}
Crystalline materials can be defined as infinite point clouds. A periodic structure can be represented as a network where a group of points is repeated by a discrete translation, which is is equivalent to parallelepiped tiling containing a cloud of atoms as illustrated in Figure \ref{fig:material_graph}. A crystal can be described as atomic positions $x_i\in[0,1[^3$ with an associated feature space $F$ representing the chemical information of each atom $z_i\in F$ and a lattice $\rho\in\text{GL}_3(\mathbb{R})$ representing the material periodicity. The infinite point cloud generated by this representation can be defined as follows:
\begin{equation}
\label{eq:pointcloud}
    \big\{ \big(\rho (x_i + \tau),\: z_i\big) \:|\: \tau \in \mathbb{Z}^3,\:1 \leq i \leq n \big\} \;\subseteq\; \mathbb{R}^3 \times F
\end{equation}

Where $\tau$ acts as a $\mathbb{Z}^3$ vector that translates the point cloud. Equation \ref{eq:pointcloud} defines the space in which the atoms are located as a torus. In fact,  when atoms leave by one side of the lattice they enter by the opposite side with the same orientation. $\text{GL}_3(\mathbb{R})$ defines the shape of the lattice, i.e\ the periodicity. $F$ is the feature space that can encode chemical information such as atomic number or charge. For crystal generation, we need to define a model capable to deform the geometry of a structure in order to minimize the total energy and hence obtain a stable structure. Such actions are performed on the material lattice $\rho$ resulting in the updated lattice $\rho'$ and on atomic positions $x_i$ resulting in the updated positions $x'_i$.
\begin{equation}
    \begin{cases}
        \rho' = h\rho       \\
        x'_i  = [x_i + h_i] \\
    \end{cases}\,.
\end{equation}
We aim to predict the action $h\in\text{GL}_3(\mathbb{R})$ on the lattice and the actions $h_i\in\mathbb{R}^3$ on the atomic position. The atomic positions are brought back into the crystal lattice by truncation. In the following, we introduce our model that learns  arbitrary deformations on crystal lattices. We first explain, in Section \ref{materials_equivariance}, why group actions are needed for materials, recall the notion of equivariance, and define our group actions on crystals while providing their properties.  Finally, Section \ref{equivariant_gnn} gives an explicit description of our model along with equivariance results. Proofs and additional materials are provided in an online ArXiv appendix.
%\footnote{\url{linktoarxiv}}.\todo{add link}

%%%%%%%%%%%%%%%%%%%%%%%%%%%%%%%%%%%%%%%%%%%%%%%%%%%%%%%%%%%%%%

\section{Equivariance and Group Actions}
\label{materials_equivariance}
Crystals materials can be seen as an infinite cloud of atoms as $\cloud{m} \subseteq \R^d\times F$. As such, equivalences between materials are defined by isometries, i.e. by the group action of $\Euc(d)$ regardless of lattice generators\footnote{A generator is a lattice property that defines pattern repetition) and atom indices}. 
As a crystal lattice can have multiple space-tiling representations resulting in an identical infinite atomic cloud, the $SL_d(\Z)$ group action is needed for paving. Consequently, the group $G = \Euc(d) \times SL_d(\Z) \times {\mathfrak S}_n$ acts on the lattice without affecting its properties. ${\mathfrak S}_n$ is the permutation group that acts by changing the numbering of atoms, where $n$ is the number of atoms. Please note that atoms are always in the same place, but not with the same index. As chirality has an impact on the properties of a chemical structure, the reflection action should be excluded. The special Euclidean group that doesn't include reflection should be then considered. However, in this work, we consider $\Euc(d)$ that acts on the chirality assuming that this limitation will not be problematic with inorganic material.
We consider crystals described by an infinite cloud of atoms
that is invariant under a discrete subgroup $L \subseteq \mathbb{R}^d$ of maximal rank. For any choice of generators $(\tau_1, \dots, \tau_d) \in L$, we consider the unique automorphism $\rho \in GL_d(\mathbb{R})$ that maps the canonical basis of $\mathbb{R}^d$ to the generating basis of $L$ to represent $L$. 
\begin{definition} 
  The representation space of {\em featured materials} ${\cal M}^F$ is the disjoint union $\coprod_{n \in \mathbb{N}} \M^F_n$ where:
    $$ {\cal M}^F_n =
        \big\{(\rho, x, z) \:|\:  \rho \in GL_d(\mathbb{R}),
        \: x \in [0, 1[^{n \times d},
        \: z \in F^n \big\} $$
    Chemical materials are represented in $\M = \M^{\mathbb{N}}$, with atomic numbers as
    feature sequence $z$.
\end{definition}

%\todo{the following definition says what ? définition 1 c'est une définition formel de l'ensemble des matériaux avec des atomes dans l'espace de feature $F$ et à $n$ atomes. C'est un ensemble infinie de triplet $\rho$, $x$, $z$ qui représente tout les matériaux à $n$ atomes possibles. Donc $\mathcal{M}^F$ c'est l'union des matériaux à 1, 2, $\cdots$ atomes, donc l'ensemble de tout les matériaux possible et imagineable}
$\M^F_n$ is an infinite set of triplet $\rho$, $x$, $z$ that represent all possible materials with $n$ atoms. The atomic number has a chemistry reference, e.g.\ 1 for hydrogen or 6 for carbon. 
\begin{definition}
    The infinite point cloud $\cloud{M}$ associated to a material $M = (\rho, x, z)$ in ${\cal M}^F_n$ is defined as:
    $$ \cloud{M} = \big\{ \big(\rho \cdot (x_i + \tau),\: z_i\big)
        \:|\: \tau \in \mathbb{Z}^d,\:1 \leq i \leq n \big\} \;\subseteq\;
        \mathbb{R}^d \times F $$
    The cloud $\cloud{M}$ is invariant under the action of the lattice $L = \rho \cdot \mathbb{Z}^d \subseteq \mathbb{R}^d$.
    \label{infinit_material}
\end{definition}

The $\Euc(d)$ group acts naturally on subsets of $\R^d$ and
two materials $M$ and $M'$ should be considered physically identical if they span isometric point clouds. 
Let us write $M \sim M'$ if there exists an isometry 
$g \in \Euc(d)$ such that $\cloud{M'} = g \cdot \cloud{M}$.
Let $\cloud{\M^F}$ be the image of $\M^F$ in ${\cal P}(\R^d \times F)$ under $\cloud{-}$. The quotient space $\M / \sim$ of equivalent materials is defined by the following universal diagram:
%Plus courant de représenter un quotient comme ça:
\[
\begin{tikzcd}
\M \rar \dar[swap]{\pi_\sim} & \cloud{\M} / \Euc(d) \\
\M / {\sim} \urar[dashed,swap]{\rm iso} & 
\end{tikzcd}
\]
Infinite point clouds can only be represented by non-intrinsic representatives $M \in \M^F$. In the following, we describe how the relation $\sim$ is related to group actions on $\M^F$. The following proposition introduces the group actions that don't change the properties of materials, i.e.\ actions that lead to producing equivalent materials.  

\begin{proposition} \label{action-Euc}
    The following four actions on $\M^F_n$ preserve the equivalence class of material:    
    \begin{itemize}
        \item $\mathfrak{S}_n$ permutation group, acting by
              $\sigma \cdot (\rho, x, z) = (\rho, x \circ \sigma^{-1}, z \circ \sigma^{-1})$
        \item $O(d)$ orthogonal group, acting by
              $g \cdot (\rho, x, z) = (g \cdot \rho, x, z)$
       % \item $\R^d$ translation group, acting by
         %     $t \cdot (\rho, x, z) = (\rho, \module{ x+t }, z)$
        \item $E$ translation group\footnote{
                  The actions of $E$ and $\R^d$ are equivalent, being simply intertwined by the isomorphism $\rho : \R^d \to E$. The action of $E$ is more natural, extending the action of $O(d)$ to $\Euc(d)$ but the action of $\R^d$ is more convenient in our representation space.
              }, acting by
              $v \cdot (\rho, x, z) = (\rho, \module{ x + \rho^{-1}v}, z)$
        \item $\Euc(E) = E \rtimes O(E)$ euclidian group, with the action induced by those of $E$ and $O(d).$  
    \end{itemize}
    These actions are free and proper on $\M^F_n$. The point cloud map $\cloud{-}$ commutes with
    these actions\footnote{
        Permutations acting trivially on $\cloud{\M}$.
    }.
\end{proposition}

Performing modification by permutations and isometries is not enough to get a faithful
representation of $\M^F / \sim$.
Different choices of lattice $L \subseteq \mathbb{R}^d$ lead to different primitive point clouds in $[0, 1[^d$. The action of $SL_d(\mathbb{Z})$ on $GL_d(\R)$ describes
all the possible choices of generators for $L$. However, $SL_d(\mathbb{Z})$
cannot simply act by left multiplication on ${\cal M}^F$ like $\Euc(d)$ without
distorting the relative positions of atoms in the primitive cell $\rho \cdot
[0, 1[^d$. We complete Proposition \ref{action-Euc} by specifying how to repave the space while being equivalent to the structure we  start with. %This is the main difference between organic and molecular chemistry.

\begin{figure}[t]
  \centering
  \includegraphics[width=0.8\columnwidth]{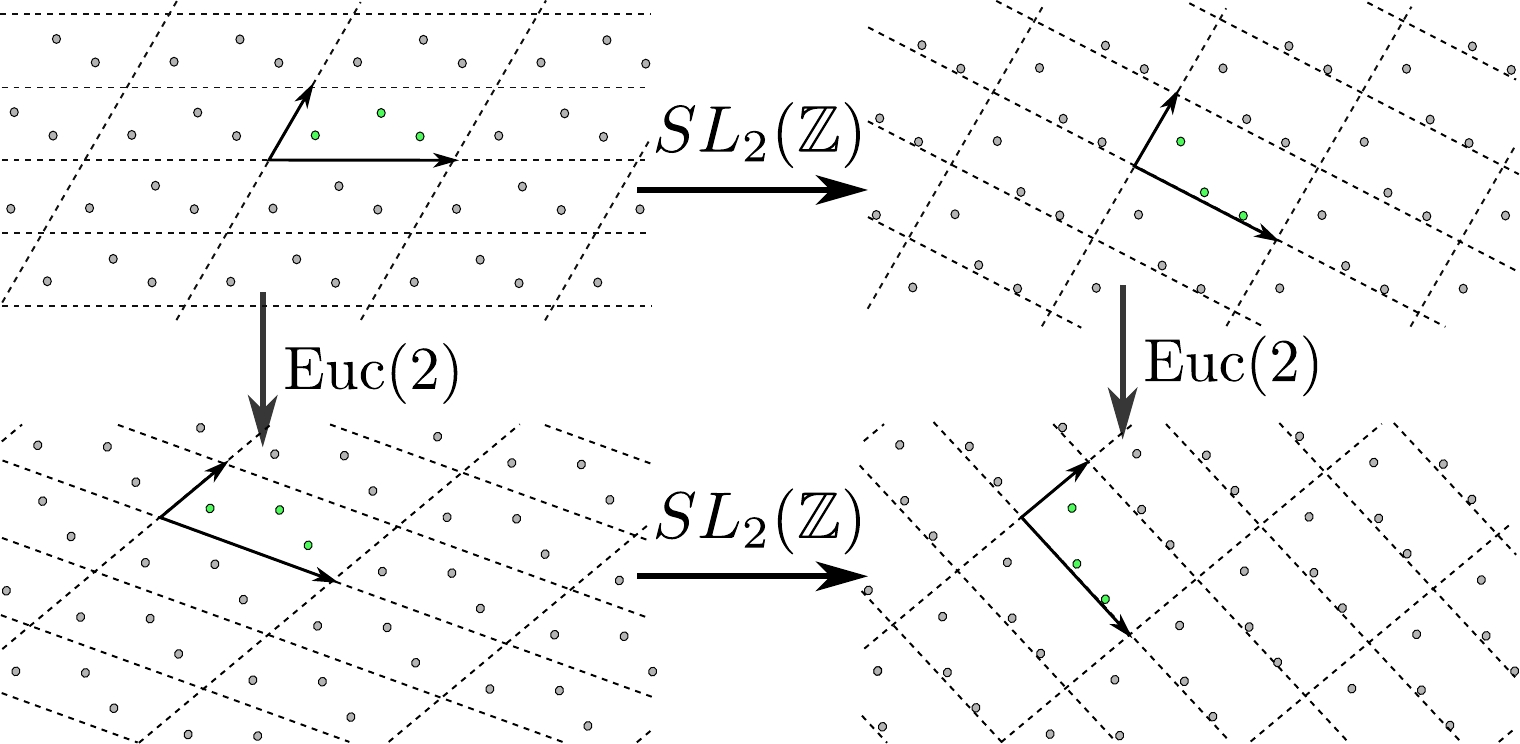}
  \caption{In definition \ref{infinit_material} the point cloud is a space tiling (top left corner). The actions from $\Euc(2)$ and $SL_2(\Z)$ groups commute and do not affect interatomic distances.}
  \label{fig:commute}
\end{figure}

\begin{proposition} \label{action-SL}
    The group $SL_d(\mathbb{Z})$ acts on $\M^F$ by letting for every change of
    lattice generators $g$:
    $$ g \cdot (\rho, x, z) = (\rho \cdot g^{-1}, \module{ g x }, z) $$
    where $\module{ g x }_i$ denotes the unique element of $[0, 1[^d$ in the
    orbit of $g x_i$ under $\mathbb{Z}^d$. Identifying the reference cell $[0,
    1[^d$ with the torus $\mathbb{T}^d$, the action of $SL_d(\mathbb{Z})$ on
    $\M^F \simeq GL_d(\mathbb{R}) \times (\mathbb{T}^d)^n \times F^n$ is free
    and proper. The point cloud map is invariant under the action of
    $SL_d(\mathbb{Z})$.
\end{proposition}

The reference cell is the base cell we use to pave the space with $L$. It is a parallelepiped of atoms and $L$ is the translation that allows the parallelepiped moving to pave the space. 
%In the following, we show that the different actions are commutative. 

\begin{proposition}
The actions of $\Euc(d)$, $\mathfrak{S}_n$ and $SL_d(\mathbb{Z})$ on
$\M^F_n$ commute as shown in Figure \ref{fig:commute}. 
%The proof of commutativity of $\Euc(d)$, $\mathfrak{S}_n$ and $SL_d(\mathbb{Z})$ on $\M^F_n$ is available in 
%supplementary material.
\end{proposition}

Let $G$ be the product of $\Euc(d) \times {\mathfrak S}_n \times
SL_d(\mathbb{Z})$.  Propositions \ref{action-Euc} and \ref{action-SL} imply
that the quotient of $\M^F$ under the action of $G$ is a well-formed
topological space. This quotient is not the space $\M^F / \sim$ of
equivalent materials, because the lattice associated with a material representation
$M \in \M^F$ is not always a maximal symmetry subgroup of its point cloud.

\paragraph{Graph equivariance} Internal forces acting on a crystal structure are equivariant to the aforementioned group actions. As the properties of a crystal depend on interatomic interaction, equivariance could be then considered as the solution to obtain generalization capability. In this work, we take advantage of the equivariance of the graph representation of materials under $G$ the product of $\Euc(d) \times {\mathfrak S}_n \times SL_d(\mathbb{Z})$. 

\begin{definition} \label{def-invariant}
    A neural network $f_\theta : \M^F \to \R^k$ is said {\em invariant} under $G$ if
    for all $g \in G$:
    $$ f_\theta(g \cdot M) = f_\theta(M) $$
\end{definition}

\begin{definition} \label{def-equivariant}
    A neural network $\ph_\theta : \M^F \to \M^{F'}$ is said {\em equivariant} under $G$ if
    for all $g \in G$:
    $$ \ph_\theta(g \cdot M) = g \cdot \ph_\theta(M) $$
\end{definition}

%%%%%%%%%%%%%%%%%%%%%%%%%%%%%%%%%%%%%%%%%%%%%%%%%%%%%%%%%%%%%%%%%%%%%%%%%%%%%%%%%%%%%%%%%%%%%%%%%%%%%%%%%%%%%%%%%%%%%%%%%%%%%%%%%%%%%%%%%%%%%%%%%%%%%%%%%%%%%%%%%%%%%%%%%%%%%%%%%%%%%%%%%%%%%%%%%%%%%%%%%%%%%%%%%%%%

\section{Equivariant GNN for Materials}\label{equivariant_gnn}

We now introduce our MPNN that performs arbitrary deformation by reasoning on relative atomic distances and angles. A spatial equivariance is enforced by the MPNN. We first associate a graph with a material and then take advantage of the local invariance (input quantities are themselves invariant: distance, angle, etc.) and equivariance of the graph to define equivariant actions on crystal lattices.

\begin{definition}
We call directed 2-graph $\Gamma = (\Gamma_0, \Gamma_1, \Gamma_2)$ 
a triplet of sets together with applications: 
\begin{itemize}
\item $\pi_1 : \Gamma_1 \to \Gamma_0 \times \Gamma_0$, written $\pi_1(\gamma) = (\src{\gamma}, \tgt{\gamma})$
\item $\pi_2 : \Gamma_2 \to \Gamma_0 \times \Gamma_0 \times \Gamma_0$ 
\end{itemize}
We call $\Gamma$ a directed 1-graph when $\Gamma_2 = \varnothing$.
\end{definition}

%\noindent{\bf Remark.}
The aforementioned graphs are often called "multi"-graphs. 
Recall that $\pi_1$ and $\pi_2$ may not be injective. 
They are called "hyper"-graphs as well, because they generalise 1-graphs
to dimensions $\geq 1$ and "directed" because we do not assume any symmetry 
on $\Gamma$ w.r.t vertice permutations. 
%\todo{introduce $\gamma$, the definition is not clear.. Gamma est un multigraphe. Donc c'est un graphe mais ça particuliarité c'est que plusieurs arêtes peuvent exister entre une paire de noeuds. Les noeuds représente des atomes et les arêtes et triplets des intéractions à 2 ou 3 corps entre atomes.}

\begin{definition}
\label{def6}
    Let $M = (\rho, x, z)$ in $\M_n^F$ be a material and $c_i > 0$
    for $1 \leq i \leq n$ denotes cutoff distances.
    We define a directed 2-graph
    $\Gamma = \Gamma_{M,c}$ by the graded components: 
    \begin{itemize} 
    \item $\Gamma_0 = \{1, \dots, n \}$ 
    \item $\Gamma_1 = \big\{ (i, j, \tau) \in \Gamma_0 \times \Gamma_0 \times
        \Z^d \: \big|\: || \rho (x_j - x_i + \tau) || < c_i \big\}$ 
    \item $\Gamma_2 = \big\{ (\gamma, \gamma') \in \Gamma_1 \times \Gamma_1
        \: \big|\: \tgt{\gamma} = \src{\gamma'} \big\}$
    \end{itemize}
    with obvious projections, i.e. with 
    $\pi_1 : (i, j, \tau) \mapsto (i, j)$ and
    $\pi_2 : (\gamma, \gamma') \mapsto (\src{\gamma}, \tgt{\gamma}, \tgt{\gamma'})$.  
\end{definition}

% \begin{figure}
%     \centering
%     \includegraphics[width=0.2\textwidth]{images/graphe_gamma_v2.pdf}
%     \caption{The graph from Definition \ref{def6} has nodes ($\Gamma_0$), edges ($\Gamma_1$) and triplets ($\Gamma_2$). Nodes represent atoms while edges and triplets represent 2 and 3 body interactions.}
%     \label{fig:graph_gamma}
% \end{figure}

This graph construction includes many definitions of material graphs, making it versatile and usable in most contexts since a material graph is built from the local environment of atoms.
This definition includes a graph built from a constant cutoff distance (i.e. $c_i$ is constant) and a graph built from $k$ nearest neighbour or built from chemical properties as the covalent radii.
Definition \ref{def6} generalizes to most of the graphs defined in previous works \cite{jorgensen2018neural,doi:10.1021/acs.chemmater.9b01294,satorras2021en}. The key feature of this construction is the invariance of edges and triplets. As interatomic distances and unoriented angles are invariants to $\Euc(d)$ and $SL_d(\Z)$ groups, any graph constructed from the local environment of the atoms will be invariant. More details about graph construction are in the appendix. 
We now introduce notations needed to define our model.

\begin{definition}
    Let consider $M = (\rho, x, z) \in \M^F$ and $\Gamma = \Gamma_{M, c}$, we introduce the following 
    notations:
    \begin{itemize} 
    \item $e_{ij}^\tau=(x_j - x_i + \tau)$ 
    for edge vector in lattice coordinates, 
    \item $v_{ij}^\tau = \rho(e_{ij}^\tau)$ for the edge vector 
    in physical space,
    \item $r_{ij}^\tau = ||v_{ij}^\tau||$ for the physical edge length, 
    \item $\theta_{ijk}^{\tau\tau'}$ as the unoriented angle between $v_{ij}^\tau$ and $v_{jk}^{\tau'}$
    \item $\mathcal{A}_{ijk}^{\tau\tau'}$ as the area of the triangle $x_i$, $x_j+\tau$ and $x_k+\tau'$
    \end{itemize}
    Let us also write $e_\gamma, v_\gamma, r_\gamma, \theta_{\gamma\gamma'}, \mathcal{A}_{ijk}^{\tau\tau'}$
    for the same quantities when we do not need to make vertices explicit.
    Note that $r_\gamma$, $\theta_{\gamma\gamma'}$ and $\mathcal{A}_{ijk}^{\tau\tau'}$ are natural Euclid invariants. 
\end{definition}

\subsection{Gradient of the invariant geometry}
\label{gradient}
To build vector fields of our equivariant MPNN, we take advantage of the gradient of the invariant geometry of crystal graphs. For 0-chains, i.e.\ vertices $i \in \Gamma_0$, the Euclid group acts 
transitively on spatial coordinates such that $I_i$ is trivial (a point) and $r_i$ is a constant. For 1-chains, i.e.\ directed edges $\gamma \in \Gamma_1$, the only Euclid invariant is the length of the associated vector. For  
$I_\gamma = \R$ and  
for $\gamma : i \overset{\tau}{\to} j$, we let:
\begin{equation} 
r_\gamma(x_\gamma) = r^\tau_{ij} 
\end{equation}
 
\noindent For 2-chains
$\cell = i \overset{\tau}{\to} j \overset{\tau'}{\to} k$, we find more convenient to define invariants as
two vector lengths and the angle at their common point, i.e.
$I_{\cell} = \R^3$ with: 
\begin{equation}
    r_{\cell} = 
    \big(\theta^{\tau\tau'}_{ijk},\,r_{ij}^\tau,\,r_{jk}^{\tau'}\big)
\end{equation}

\noindent For a tangent vector at $\rho \in GL_d(\R)$, we have:
\begin{equation}
\frac{\partial v_{ij}^\tau}{\partial \rho} = 
\rho \cdot (x_j - x_i + \tau) 
= \rho \cdot e_{ij}^\tau 
\end{equation}
The differential edge distances with respect to $\rho$ projects on the source and image edge vectors $e_{ij}^\tau$ and $u_{ij}^\tau$ respectively. It is equal to 1 on the rank 1 linear map $|u_{ij}^\tau\rangle \langle e_{ij}^\tau|$. $u_{ij}^\tau$ denotes the normalized vector $v_{ij}^\tau$ such as $u_{ij}^\tau = v_{ij}^\tau / r_{ij}^\tau$.
\begin{equation}
\frac{\partial r_{ij}^\tau}{\partial \rho}  
= \langle u_{ij}^\tau \, ,\, \rho \cdot e_{ij}^\tau \rangle
\end{equation}
The angle differentials with respect to $\rho$ are computed by assuming that the middle point is fixed (it is true up to a translation in the target space, which does not alter the angle). $\omega_{ijk}^{\tau\tau'}$ denotes the unit normal vector to $(v_{ij}^\tau, v_{jk}^\tau)$
\begin{equation}
\frac{\partial \theta_{ijk}^{\tau\tau'}}{\partial \rho} 
= \langle \omega_{ijk}^{\tau\tau'} \times u_{jk}^{\tau'} 
\rho \cdot e_{jk}^{\tau'} \rangle 
-  \langle \omega_{ijk}^{\tau\tau'} \times u_{ij}^\tau 
\rho \cdot e_{ij}^\tau \rangle
\end{equation}
%{\bf Remark.} 
The mixed product coincides with the determinant and is invariant under cyclic permutations.

\subsection{Equivariant Message Passing Neural Network}
We now introduce a general definition of our equivariant MPNN based on vector fields.  We formally define $\lambda$ as the vector field used in Equation\ref{gnn_rho}. It allows for defining how the GNN acts on the crystal lattice. 

\begin{definition}\label{forcefields}
To every edge $\gamma \in \Gamma_1$ and every 2-region 
$\gamma\gamma' \in \Gamma_2$ we associate 
the infinitesimal lattice deformations
$\lambda_\cell : \M_\cell \to \mathfrak{gl}_d$ 
defined by: 
\begin{itemize}
    \item $\lambda_{\gamma}(M_\cell) = \ket{u_\gamma} \bra{u_\gamma}$ 
    \item $\lambda_{\gamma\gamma'}(M_\cell) = \ket{u_\gamma} \bra{u_{\gamma'}} + \ket{u_{\gamma'}} \bra{u_\gamma}$ 
\end{itemize}

The $\ket-\bra$ is a notation in quantum physics to denote the matrix obtained as the product of a column vector ($\ket(V)$ is $V$ seen as a column) and a line vector ($\bra W$ is W seen as line vector). In our case $\ket-\bra$ with two vectors $u,v\in\R^d$ we have $\ket{u}\bra{v}=uv^\intercal$. 
Alternatively, we can directly use gradients of the geometric invariant such as:
\begin{itemize}
    \item $\lambda_{\gamma}(M_\cell) = \frac{\partial r_{ij}^{\tau}}{\partial \rho} $ 
    \item $\lambda_{\gamma\gamma'}(M_\cell) = \frac{\partial r_{ij}^{\tau}}{\partial \rho} \text{ or } \frac{\partial r_{ik}^{\tau'}}{\partial \rho} \text{ or } \frac{\partial \theta_{ijk}^{\tau\tau'}}{\partial \rho} \text{ or } \frac{\partial \mathcal{A}_{ijk}^{\tau\tau'}}{\partial \rho} $ 
\end{itemize}
To ensure transversality with $\mathfrak{so}_d$, $\lambda_{\cell}$ for all $\cell \in \Gamma$ is symmetric as equivariance means that the lattice is searched among an equivalence class in $GL_d(\R) / SO_d$.
\end{definition}

An equivariant GNN that acts on materials is as follows:

\begin{proposition} \label{prop-equiv}
    A neural network
    $\ph_\theta : {\cal M}_n^F \to {\cal M}_n^{F'}$,
    written $\ph_\theta : (\rho, x, z) \mapsto \rho' $
    is decomposed as follows:

    The generation of messages from the edges and the triplets of the graph such as $\ph_\theta^{m^{(k)}}: \R^{f^{(k)}\times\Gamma_k} \to \R^{h^{(k)}\times\Gamma_k}$
    \begin{subequations}
        \begin{align}
            m_{ij\tau} & = \ph_\theta^{m^{(1)}}(z_i, z_j, ||v_{ij\tau} ||)\\
            m_{\gamma,\gamma'} & = \ph_\theta^{m^{(2)}}(z_i, z_j, z_k, ||v_\gamma ||, ||v_{\gamma'}||,\theta_{\gamma,\gamma'})\label{gnn_message}
        \end{align}
    \end{subequations}
    The aggregation and update of the messages at each node is $\ph_\theta^{z^{(k)}}: \R^{h^{(k)}\times\Gamma_k} \to \R^{h'^{(k)}\times\Gamma_k}$ and $\ph_\theta^u: \R^{z\times\Gamma_0}\times\R^{h'^{(1)}\times\Gamma_1}\times\R^{h'^{(2)}\times\Gamma_2} \to \R^{z\times\Gamma_0}$
    \begin{align}
        z'_i = & \ph_\theta^u(z_i, \sum_{\gamma \in \Gamma_1(i)} \ph_\theta^{z^{(1)}}(m_\gamma), \sum_{(\gamma,\gamma') \in \Gamma_2(i)} \ph_\theta^{z^{(2)}}(m_{\gamma\gamma'}))
        \label{gnn_update}
    \end{align}
    $\ph_\theta^{\rho^{(k)}}$ is the weight of a vector field $\lambda_{\cell}$ such as $\ph_\theta^{\rho^{(k)}}:\R^{f'^{(k)}\times\Gamma_k} \to \R^{\Gamma_k}$
    \begin{subequations}\label{gnn_rho}
        \begin{align}
            \rho' & = \exp\left(\frac{1}{|\Gamma_1|}\sum_{\gamma \in \Gamma_1} \ph_\theta^{\rho^{(1)}}(m_\gamma) \cdot \lambda_{\gamma}\right) \cdot \rho\label{gnn_rho_edges}\\
            \rho' & = \exp\left(\frac{1}{|\Gamma_2|}\sum_{(\gamma,\gamma') \in \Gamma_2} \ph_\theta^{\rho^{(2)}}(m_\gamma,m_{\gamma'},\theta_{\gamma\gamma'}) \cdot \lambda_{\gamma\gamma'}\right) \cdot \rho\label{gnn_rho_triplets}
        \end{align}
    \end{subequations}
    $\ph_\theta$ is equivariant under $G = \Euc(d) \times {\mathfrak S}_n \times SL_d(\mathbb{Z})$ 
    if the vector field $\lambda_\cell$ is invariant to $SL_d(\Z)$ and equivariant to $\Euc(d)$ such as $\lambda_\cell(g \cdot M) = 
    g \lambda_\cell(M) g^{-1}$ for all $g\in O(d)$, as the translation doesn't act on the crystal lattice.
\end{proposition}

From proposition \ref{prop-equiv}, a GNN architecture acting on crystal material that satisfies Equations 8-10 is equivariant.    

\subsection{EMPNN for Crystal Lattice Deformation}
\label{EMPNN}
\begin{figure}
\centering
\begin{subfigure}[b]{0.45\linewidth}
  \centering
  \includegraphics[width=0.8\textwidth]{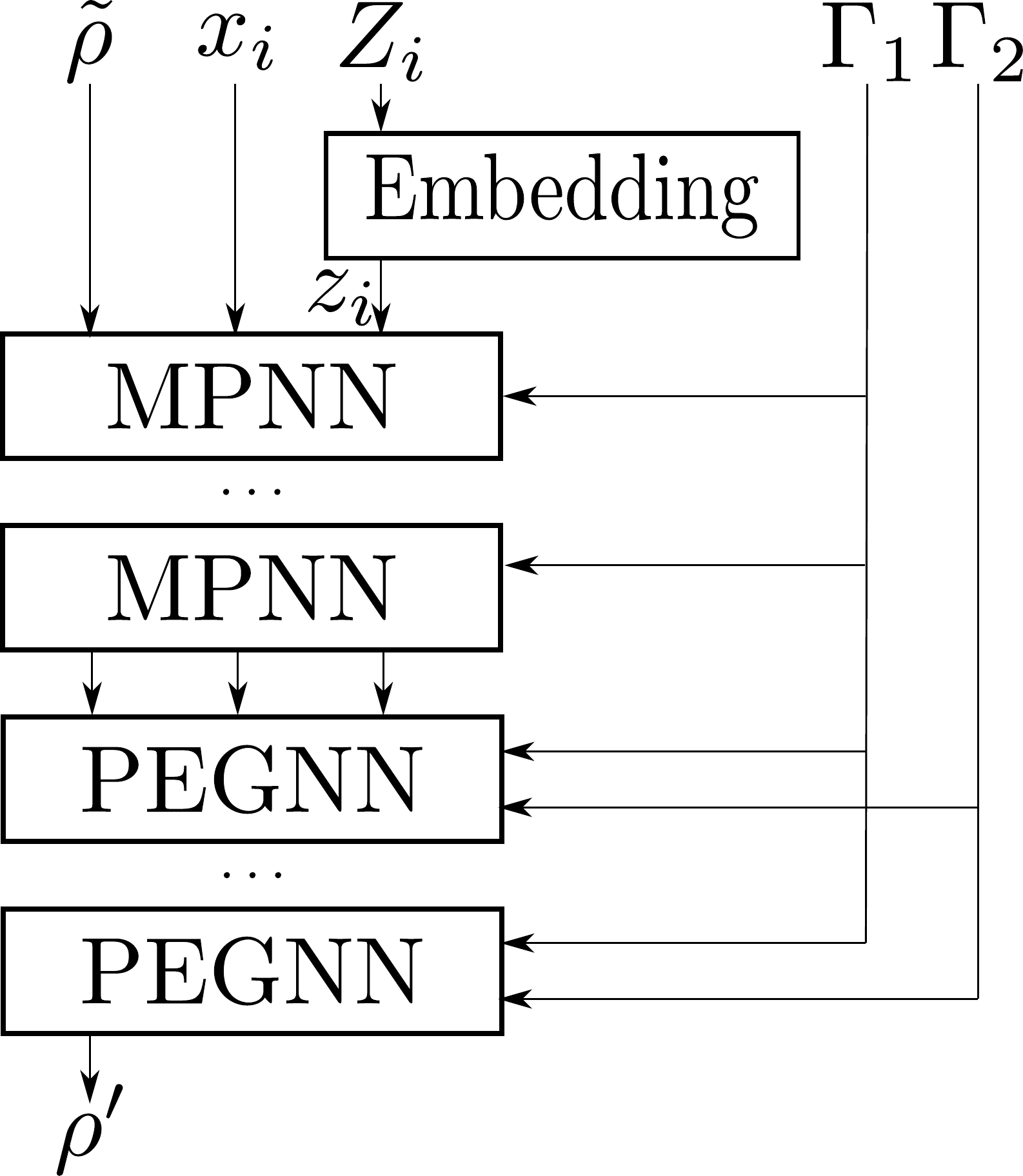}
  \caption{Overview of our model}
\end{subfigure} 
\hfill
\begin{subfigure}[b]{0.45\linewidth}
  \centering
  \includegraphics[width=0.9\textwidth]{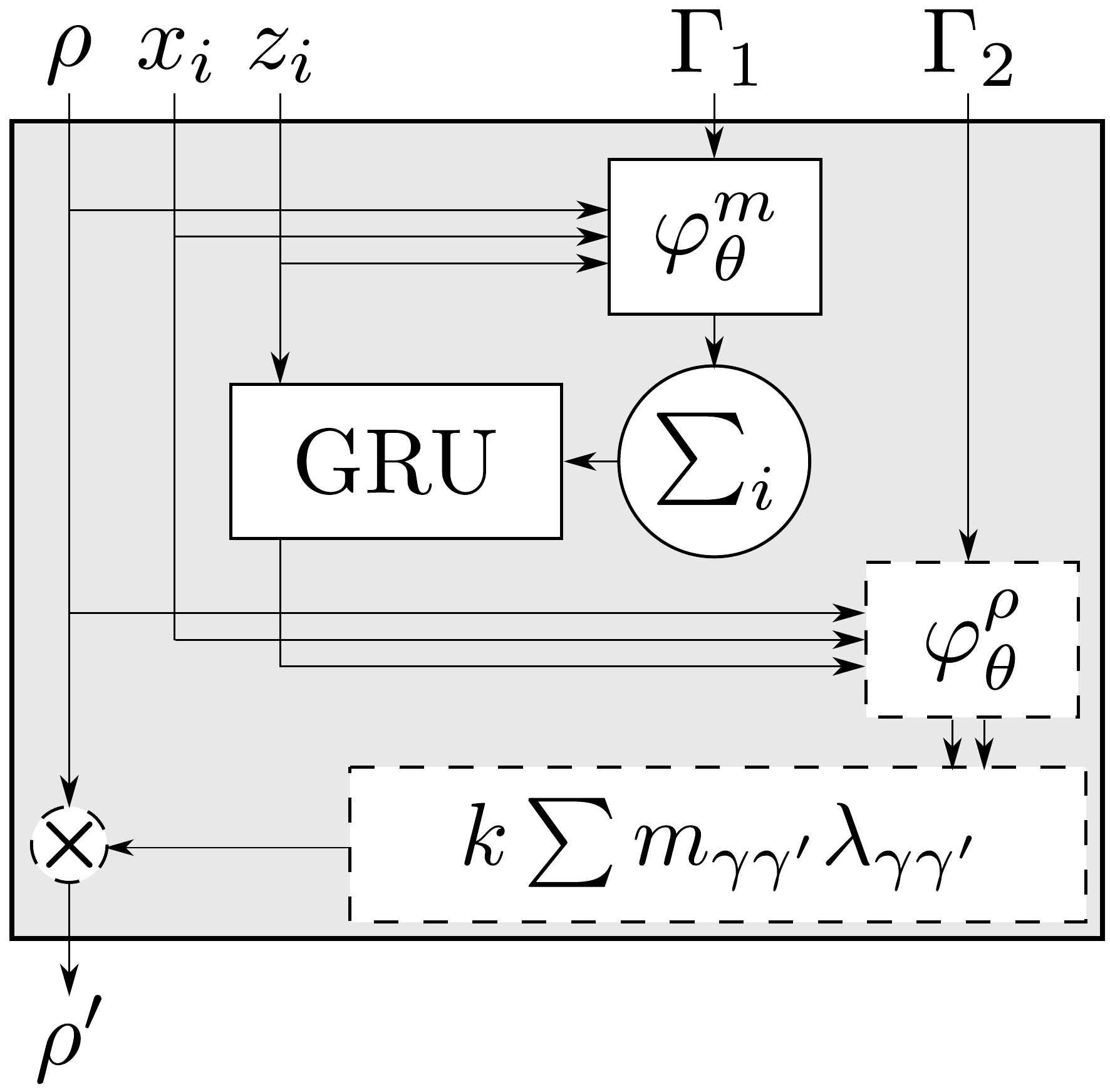}
  \caption{EMPNN layer}
\end{subfigure}
\small\caption{(a) The EMPNN model comprises an embedding layer, standard MPNN layers and EMPNN layers to perform deformation. (b) A EMPNN layer is composed of an MPNN with vector fields deforming the lattice $\rho$.}
\label{gnn}
\end{figure}

% \todo[inline]{We now introduce our GNN architecture that acts on the lattice of a crystal. 
% %Our model consists of a classical MPNN with a basic propagation scheme over the edges of the 1-graph.
% %\subsubsection{Message Passing Neural Network}
% Our periodic equivariant MPNN layer takes the lattice of the crystal $\rho^l\in\text{GL}_3(\mathbb{R})$, a set of embedding $h^l=\{h^l_1,\dots,h^l_n\}$, a set of atomic position $x^l=\{x^l_1,\dots,x^l_n\}$, edges $\gamma\in\Gamma_1$ and triplets $(\gamma,\gamma')\in\Gamma_2$ from the k-nearest neighbour graph. The triplets are formed from the edges when two edges share a common node inside of the crystal cell. The message passing equations for the layer $l$ are defined as follows:}
To empirically evaluate our approach, we defined EMPNN as a simple but effective GNN model that fits with Proposition \ref{prop-equiv}. We chose to keep our model simple to facilitate the comparison between multiple vector fields. The architecture is illustrated in Figure \ref{gnn}. We slightly adapted equation \ref{gnn_rho} by adding a first-order approximation of the matrix exponential to both vector fields over the edges and the triplets. Further details are given in Section B.2 of the appendix.

\subsubsection{Loss functions}
The goal of a loss function is to reproduce the shape and volume of the target crystal, i.e.\ $\ph(\tilde{\rho}\cdot h^{-1})=g\cdot \rho\cdot h^{-1},g\in O(3) \text{ and } h\in SL_3(\Z)$ (as $\Euc(3)$ acts on $\rho$ as $O(3)$).  There exist multiple ways to define loss functions, but all the definitions will have implicit bias. To evaluate this bias, we use a classical loss function over the normalized lattice parameters. Another approach is to compute a matrix distance between the metric tensors. Both the lattice parameters and metric tensor losses are invariant to the euclidean group but equivariant to $SL_3(\Z)$. We tested the mean absolute error (MAE) $\mathcal{L}^\text{Param}_\text{mae}$ and the mean squared error (MSE) $\mathcal{L}^\text{Param}_\text{mse}$ of the normalized lattice parameters. We have also tested the MAE $\mathcal{L}^\rho_\text{mae}$, the MSE $\mathcal{L}^\rho_\text{mse}$ and the invariant Riemannian metric $\mathcal{L}^\rho_\text{Riemann}$ of the metric tensors. The loss expressions are available in the appendix.

%%%%%%%%%%%%%%%%%%%%%%%%%%%%%%%%%%%%%%%%%%%%%%%%%%%%%%%%%%%%%%%%%%%%%%%%%%%%%%

\section{Experiments}

\begin{table*}[t]
    \small
    \centering
    \begin{tabular}{l|ccc|ccc|ccc}
        \multicolumn{1}{c|}{\multirow{2}{*}{loss}} & \multicolumn{3}{c|}{Carbon-24} & \multicolumn{3}{c|}{Mp-20} & \multicolumn{3}{c}{Perov-5}\\
    & lengths & angle & energy & lengths & angle & energy & lengths & angle & energy\\
    \hline
    $\mathcal{L}^\text{Param}_\text{mae}$ & \textbf{0.696} & \textbf{8.390} & \textbf{-0.655} (62.5) & \textbf{0.785} & \textbf{5.093} & \textbf{3.124} (51.7) & \textbf{0.967} & \textbf{15.227} & -3.426 (93.8)\\
    $\mathcal{L}^\text{Param}_\text{mse}$ & 0.677 & 8.148
    & -0.413 (65.6) & 0.710 & 4.752 & 5.485 (44.8) & 0.983 & 15.437 & -3.634 (93.8)\\
    $\mathcal{L}^\rho_\text{mae}$ & 0.599 & 4.306 & 0.526 (62.5) & 0.540 & 1.674 & 11.268 (40.7) & 0.964 & 15.074 & -1.518 (90.6)\\
    $\mathcal{L}^\rho_\text{mse}$ & 0.655 & 5.563 & 2.432 (40.6) & 0.683 & 2.645 & 10.964 (18.5) & 0.974 & 15.047 & \textbf{-3.741} (93.8)\\
    $\mathcal{L}^\rho_\text{Riemann}$ & 0.637 & 5.352 & 0.864 (43.8) & 0.729 & 3.777 & 6.859 (51.7) & 0.967 & 15.367 & -3.088 (93.8)\\
    \end{tabular}
    \caption{Metrics are defined as the average improvement of lattice parameters and the average improvement of total energy. The metrics are calculated between noisy structure and denoised structure. The lengths are given in {\AA} (angström), the angle in degree (higher is better) and the energy in eV/atom (lower is better). The value between parenthesis is the percentage of structure with lower energy. Energy is calculated with VASP\cite{PhysRevB.47.558,KRESSE199615,PhysRevB.54.11169} on a subset of 32 structures because of the high computational budget of DFT calculation.}
    \label{loss_benchmark}
\end{table*}

Our main goal is to show the capability of our EMPNN to perform arbitrary crystal lattice deformation by improving the total energy of crystal structures, i.e. the thermodynamic stability.  We rely on denoising of the crystal lattice as  evaluation task\footnote{Code and data are available at \url{https://github.com/aklipf/pegnn}}. We considered datasets of stable crystals where each structure is in local minima of formation energy. Applying a small random deformation to a structure leads to a less stable one with a high energy level (as the energy increases in all directions locally). We can then generate pairs of stable and less stable structures that we used to teach our model how to deform the less stable structure to obtain a stable one. 
%As such, the denoising task can be used to assess the capability of our model to learn stable points of the energy function. 
In general, denoising tasks are more insightful than generative tasks as they show how a model acts on a crystal lattice. 
More specifically, external bias can be better controlled when performing denoising. The chemical composition and atomic positions have an important impact on the outcome. For example, binary and ternary compounds with a light element are known to be significantly easier to generate than ternary compounds without light elements or quaternary compounds. Consequently, a generator may tend to produce simple stable materials instead of a representative sample. In this case, an improvement of the metrics may not reflect lattice improvement. The quality of a crystal is also more difficult to evaluate. Namely, if a generative model can not produce some specific lattice shapes, quantitative metrics will struggle to measure the bias.
%whereas a relaxation task can do on every shape of the testing set. Also, 
%The success of generative tasks is not only conditioned by the capability to act on the lattice 
Therefore, the performance of a generator is not a good measure to evaluate the performance of  our model on arbitrary lattice deformation.
%Generative tasks are also harder to set up, limiting the number of configurations of the model that can be tested with the same computational cost.

\paragraph{Evaluation metrics} We introduce three evaluation metrics defined as the average improvement of lattice parameters and the total energy.  Let us denote the lattice parameters by $abc\in{\R}^3$ and $\alpha\beta\gamma\in{\R}^3$ and the total energy by $E\in\R$. Given a parameter $y$, let $\tilde{y}$ be the noisy parameter and  $y'$ the denoised parameter. The  metrics are defined as follows:
\begin{align}
    \text{length}=&\frac{1}{3N}\sum^N_{k=1}l1(\widetilde{abc}_k,abc_k)-l1(abc'_k,abc_k)\\
% \end{align}
% \begin{align}
    \text{angle}=&\frac{1}{3N}\sum^N_{k=1}l1(\widetilde{\alpha\beta\gamma}_k,\alpha\beta\gamma_k)-l1(\alpha\beta\gamma'_k,\alpha\beta\gamma_k)\\
% \end{align}
% \begin{align}
    \text{energy}=&\frac{1}{N}\sum^N_{k=1}E'_k-\tilde{E}_k
\end{align}
Namely, the improvement could be geometrical, i.e.\ based on lattice parameters or chemical, i.e.\  lowering of the formation energy. Evaluating the formation energy is computationally expressive and only done on a small subset of the test set.

\paragraph{Experimental setting and datasets}  
We considered three datasets of stable crystals for which we perform denoising: Perov-5 \cite{castelli2012new,castelli2012computational}, Carbon-24 \cite{carbon2020data} and Mp-20 \cite{jain2013commentary}. Perov-5 contains perovskite (cubic) structures that have highly uniform shapes but with different chemical compositions between structures. Carbon-24 is composed of carbon atoms having a large variety of shapes. This dataset is used to evaluate the performance of our EMPNN without negative bias in case of a poor chemical encoding of atoms. Mp-20 is a subset of the material project proposed in \cite{xie2021crystal} that has a large sample of shapes and chemical compositions. It is the most representative of ordinary structures. We used the same training, validation and test splits as \cite{xie2021crystal}. To train our model, we apply random deformations on the lattices $\rho$ as $\tilde{\rho}=\exp(A)\rho$ with $A\sim\mathcal{N}(0,\sigma)$. All the conducted experiments, use grid search on hyperparameters. More information about the experiments is given in the supplementary materials. We conducted three experiments to evaluate (1) the loss functions of Section \ref{EMPNN}, (2) the vector fields and (3) the reconstruction capability of our model.

\paragraph{Loss functions evaluation}
Table \ref{loss_benchmark} shows the relationship between geometrical and chemical metrics. The best energy improvements are generally associated with the best lattice error improvement. Based on lattice parameters comparison, we obtain better performance of the loss functions. However, we may expect that this evaluation is biased. But since energy-based metrics show similar results to geometry based metrics, we conclude that the bias is negligible. 
%The experiments performed a loss function using a standard deviation $\sigma=0.3$ for lattice deformation.

\begin{table*}[t]
    \small
    \centering
    \begin{tabular}{cl|cc|cc|cc}
        \multicolumn{2}{c|}{\multirow{2}{*}{method}} & \multicolumn{2}{c|}{Carbon-24} & \multicolumn{2}{c|}{Mp-20} & \multicolumn{2}{c}{Perov-5}\\
        && lengths & angle & lengths & angle & lengths & angle\\
        \hline
        \multirow{5}{*}{$|\bullet\rangle\langle\bullet|$} & $\{|\bullet\rangle\langle \bullet|\subseteq\Gamma_1\}$ & \textbf{0.084} & \textbf{1.266} & \textbf{0.115} & \textbf{1.437} & 0.290 & 5.487\\
        & $\{|\gamma\rangle\langle \gamma|\subseteq\Gamma_2\}$ & 0.056 & 0.596 & 0.053 & 0.283 & 0.287 & 5.209\\
        & $\{|\bullet\rangle\langle \bullet|\subseteq\Gamma_1\}\cup\{|\gamma\rangle\langle \gamma'|\subseteq\Gamma_2\}$ & 0.063 & 0.454 & 0.063 & 0.270 & \textbf{0.296} & 5.733\\
        & $\{|\bullet\rangle\langle \bullet|\subseteq\Gamma_1\}\cup\{|\gamma\rangle\langle \gamma|,|\gamma\rangle\langle \gamma'|_\text{sym}\subseteq\Gamma_2\}$ & 0.065 & 0.670 & 0.066 & 0.353 & \textbf{0.296} & 5.733\\
        & $\{|\bullet\rangle\langle \bullet|\subseteq\Gamma_1\}\cup\{|\bullet\rangle\langle \bullet|\subseteq\Gamma_2\}$ & 0.065 & 0.725 & 0.066 & 0.420 & \textbf{0.296} & \textbf{5.765}\\
        \hline
        \multirow{6}{*}{$\nabla$} & $\{r_\gamma\subseteq\Gamma_1\}$ & 0.075 & 1.183 & 0.102 & \textbf{1.479} & 0.259 & 4.654\\
        & $\{r_\gamma, r_{\gamma'}\subseteq\Gamma_2\}$ & 0.060 & 0.488 & 0.085 & 0.391 & 0.289 & \textbf{5.560}\\
        & $\{r_\gamma\subseteq\Gamma_1\}\cup\{r_\gamma, r_{\gamma'}\subseteq\Gamma_2\}$ & 0.101 & 1.232 & 0.101 & 0.541 & 0.292 & 5.514\\
        & $\{r_\gamma\subseteq\Gamma_1\}\cup\{r_\gamma,r_{\gamma'},\mathcal{A}_{\gamma\gamma'}\subseteq\Gamma_2\}$ & 0.087 & 1.093 & \textbf{0.106} & 0.717 & 0.265 & 4.990\\
        & $\{r_\gamma\subseteq\Gamma_1\}\cup\{r_\gamma,r_{\gamma'},\theta_{\gamma\gamma'}\subseteq\Gamma_2\}$ & \textbf{0.107} & \textbf{1.283} & 0.088 & 0.617 & \textbf{0.293} & 5.550\\
        \hline
        \multirow{6}{*}{$\nabla_\text{sym}$} & $\{r_\gamma\subseteq\Gamma_1\}$ & 0.083 & 1.307 & 0.064 & 0.816 & 0.281 & 5.134\\
        & $\{r_\gamma, r_{\gamma'}\subseteq\Gamma_2\}$ & \textbf{0.100} & 1.188 & 0.101 & 0.503 & 0.281 & 4.959\\
        & $\{r_\gamma\subseteq\Gamma_1\}\cup\{r_\gamma, r_{\gamma'}\subseteq\Gamma_2\}$ & 0.097 & \textbf{1.375} & 0.098 & 0.672 & 0.226 & 3.188\\
        & $\{r_\gamma\subseteq\Gamma_1\}\cup\{r_\gamma, r_{\gamma'},\mathcal{A}_{\gamma\gamma'}\subseteq\Gamma_2\}$ & 0.099 & 1.328 & \textbf{0.124} & \textbf{1.160} & 0.285 & 5.457\\
        & $\{r_\gamma\subseteq\Gamma_1\}\cup\{r_\gamma, r_{\gamma'},\theta_{\gamma\gamma'}\subseteq\Gamma_2\}$ & \textbf{0.100} & 1.289 & -0.001 & -0.007 & \textbf{0.291} & \textbf{5.617}\\
        \hline
        \multicolumn{2}{l|}{feed forward (FF)} & -0.191 & -5.277 & -0.304 & -3.304 & 0.303 & 6.438\\
        \hline
        \multicolumn{2}{l|}{DFT} & 0.164 & 5.442 & 0.345 & 5.648 & 0.150 & -1.446\\
    \end{tabular}
    \caption{Metrics are defined as the average improvement of lattice parameters. The experiment is split into five categories of vector fields: from the ket-bra $|\bullet\rangle\langle\bullet|$, from the gradient of invariant geometric without symmetric action $\nabla$, the gradient with symmetric action $\nabla_\text{sym}$, lattice predicted by a FF readout function and lattice obtained from a DFT calculation with VASP.}
    \label{forcefields_benchmark}
\end{table*}

\paragraph{Force field evaluation}
We evaluated force field configurations acting on the lattice. We first considered the edge information: $\{|\gamma\rangle\langle \gamma|\subseteq\Gamma_1\}$ and $\{r_\gamma\subseteq\Gamma_1\}$. Second, we consider triplets information without angle and area: $\{|\gamma\rangle\langle \gamma|\subseteq\Gamma_2\}$ and $\{r_\gamma,r_{\gamma'}\subseteq\Gamma_2\}$. As geometrical information such as angles can determine crystal properties, we include triplets information as unoriented angles and area: $\{|\gamma\rangle\langle \gamma'|\subseteq\Gamma_2\}$ and $\{|\gamma\rangle\langle \gamma|,|\gamma\rangle\langle \gamma'|_\text{sym}\subseteq\Gamma_2\}$. $\cup$, represents the union of several vector fields and $\bullet$ denotes a wildcard that takes all vector fields into account for a given n-graph. We also evaluate the benefits of symmetric matrices on the lattice as suggested in Definition \ref{forcefields}. 
Any matrix in $GL_d(\mathbb{R})$ can be seen as the composition of a rotation and a symmetric matrix, i.e.\ polar decomposition such as $M=RS$ with $R\in SO_d$ and $S\in GL_d(\mathbb{R})/SO_d$.
As rotation doesn't act on material properties, then acting on the lattice with $M$ is equivalent to acting on the lattice with a symmetric matrix $S$. Forcing this action to be a symmetric matrix may then lead to interesting results. We conduct experiments  with relaxed symmetry constraint "sym" when the symmetric vector fields are used.
%To evaluate the performance of all the conducted experiments, use grid search on hyperparameters. More information about hyperparameters is given in the supplementary materials.
%Finally, we test multiple configuration of the equivariant MPNN on the three dataset with the loss function defined as a mean average error over normalized lattice parameters because it generally performs better. 

As baselines, we first consider the (Feed-forward FF) method proposed in \cite{xie2021crystal} which is  an invariant method aiming to predict lattice parameters (distances and angles) using an invariant encoder with a simple FF. 
%This method ensures the invariance of the lattice parameters to euclidean group actions. 
%Indeed, the lattice parameters are composed of distances and angles and are preserved. 
%However, actions from the $SL_d(\Z)$ group still act on lattice parameters.
This allows us to compare the performance of our model with an invariant model. The second baseline (DFT) is a DFT calculation that evaluates the stress tensor of the crystal and optimizes its geometry. The configuration of the DFT calculation is given in the supplementary materials. DFT is not based on ML, as such, it is  computationally heavy compared to EMPNN. DFT is unsuited for generating crystals without optimization  techniques. So it cannot really be compared with ML models (baselines and our model), but we chose to use it to provide insight into the metrics.

%Table \ref{forcefields_benchmark} contains the best metric obtained during the grid search optimization. A standard deviation of $\sigma=0.1$ is used to deform the lattice. This value is smaller than on the loss function experiment but 

Table \ref{forcefields_benchmark} shows an enhanced denoising capability of our model for most of the proposed variants. Including triplets information improves the results when vector fields are defined from the gradient of invariant geometry (Section \ref{gradient}). However, vector fields defined from edges information achieved more consistent results than those defined from triplets, especially on ket-bra. Our model outperforms FF on Carbon-24 and Mp-20 with a significant improvement of the lattice parameters but not on Perov-5 (although the performance is very close).  This suggests the importance of equivariance. The FF is not capable to achieve fine-grained deformation contrary based on vector fields. In fact, FF converges much faster during the first training steps but cannot improve the loss above a certain threshold. The only case where FF outperforms our model is when the crystal shape is extremely uniform, which is the case of Perov-5 where all the structures are cubic. In Perov-5, the angle improvement is not relevant as FF uses normalized lattice parameters. A random model or a constant parameter will produce similar results. 
Regarding DFT, it improves the lattice parameters on Carbon-24 and Mp-20 but not on Perov-5. This suggests that the crystals before random deformation probably remain close to local minima of the formation energy after deformation on Carbon-24 and Mp-20, but not on Perov-5. Our methods can take advantage of the biased distribution on Perov-5 while DFT is not capable of. 
Finally, comparing multiple configurations of vector fields shows that ket-bra works better on 1-graph while gradient-based vector fields work better on 2-graph. Triplets vector fields obtain better results with the area and angle information. 

\noindent\textbf{Reconstruction task evaluation.} The reconstruction is close to the generative task and aims to build a crystal lattice from scratch. This cannot be performed with chemical simulation techniques such as DFT. We start from the point cloud as if it was in a cubic lattice of one {\AA}  on a side. From this cubic lattice, the EMPNN performs the reconstruction. The main hypothesis is that there is a single stable cell which corresponds to the starting atomic positions. Our model consistently outperforms the FF model as shown in Table \ref{reconstraction}. 
\begin{table}
    \centering
    \begin{tabular}{ c| c c| c c }
        model & \multicolumn{2}{c|}{carbon-24}  &\multicolumn{2}{c}{mp-20} \\ 
        & length & angles & length & angles \\ 
        \hline
        EMPNN & \textbf{0.200} & \textbf{3.199} & \textbf{0.174} & \textbf{1.965} \\  
        baseline & 0.469 & 13.693 & 0.534 & 6.324
    \end{tabular}
    \caption{MAE between lattice parameters of the original cell and the reconstructed cell ({\AA} and degree).}
    \label{reconstraction}
\end{table}
% \todo{add a discussion about the new resultas. Give a caption tho the table.

% Caption: )

% Discussion: Notre méthode est largement meilleurs que la baseline (mae=> plus petit est le meilleurs). On peut voir qu'en réalité les performances de la baseline en reconstruction sont assez mauvaise au niveau angulaire (13 degrée en moyenne c'est vraiment beaucoup) et pas très bonne niveau distance (l'erreur est proche de la taille d'un atome). Pourtant notre baseline est une technique très classique quand on cherche à trouver les paramètres de maille. La précision des dimensions de la maille est aussi bien meilleur avec une division par 2 ou 3 de la mae. 
% }

%%%%%%%%%%%%%%%%%%%%%%%%%%%%%%%%%%%%%%%%%%%
\section{Conclusion}
We proposed a general equivariant MPNN framework for material science by taking into consideration $SL_3(\Z)$ group action on crystal materials. In particular, our model uses multiple vector fields to act on crystal lattices. We showed the benefits of our model compared to equivariant baselines that do not consider $SL_3(\Z)$. We also compared different loss functions and results with DFT calculation to give insight into methods based on lattice reconstruction such as those using auto-encoder.
%Our model showed interesting results on multiple datasets but further investigation is required to apply our framework to generative tasks. 

\section{Acknowledgments}
This work has been supported by ANR-22-CE23-0002 ERIANA, ANR-20-THIA-0004 and by HPC resources from GENCI-IDRIS (Grant 2022-[AD011013338]).

\bibliography{aaai23.bib}

\newpage
\appendix

\section{Equivariance and Group Actions}
    
\begin{proposition}
    $\module{ \module{ a } + b }=\module{ a + b }$ with $a,b\in\R^d$.
\end{proposition}

\begin{proof}
    \begin{align*}
        \module{ \module{ a } + b }&=\module{ \module{ a } + \module{ b } + t } \text{ , with }t=b-\module{ b }\\
        &=\module{ \module{ a } + \module{ b } } \text{ as } t\in\Z^d\\
        &=\module{ a - t' + b - t" } \text{ with } t'=\module{ a }-a,t"=\module{ b }-b\\
        &=\module{ a + b } \text{ as } (t' + t")\in\Z^d.
    \end{align*}
\end{proof}

\begin{proposition}
$\module{ g\cdot\module{ x } }=\module{ g\cdot x }$ with $x\in\R^d$ and $g\in\Z^{d\times d}$.
\end{proposition}

\begin{proof}
    \begin{align*}
        \module{ g\cdot\module{ x } }&=\module{ g\cdot(x-t)}\text{ with }t=x-\module{ x}\\
        &=\module{ g\cdot x - g\cdot t }\\
        &=\module{ g\cdot x  }\text{ as }g\cdot t\in\Z^d.
    \end{align*}
\end{proof}

\subsubsection*{Commutativity of $\Euc(d)$, $\mathfrak{S}_n$ and $SL_d(\mathbb{Z})$ on $\M^F_n$}

\begin{proof}\label{proof_commutativity}
    To prove the commutativity of $\Euc(d)$ and $SL_d(\Z)$, we must show that $g\cdot g'\cdot(\rho,x,z)=g'\cdot g\cdot(\rho,x,z)$ with $g\in\Euc(d)$ and $g'\in SL_d(\Z)$. Let's see how $g$ and $g'$ act on $(\rho,x,z)$\\
    \begin{align*}
    &g\cdot g'\cdot(\rho,x,z)\\
    =&g\cdot(\rho\cdot g'^{-1},\module{ g'x},z)\\
    =&(g\cdot(\rho\cdot g'^{-1}),\module{\module{ g'\cdot x}+(\rho\cdot g'^{-1})^{-1}\cdot v},z)
    \end{align*}
    \begin{align*}
    &g'\cdot g\cdot(\rho,x,z)\\
    =&g'\cdot(g\cdot\rho,\module{ x+\rho^{-1}\cdot v},z)\\
    =&((g\cdot\rho)\cdot g'^{-1},\module{ g'\cdot\module{ x+\rho^{-1}\cdot v}},z)
    \end{align*}
    To prove the commutativity of the actions on $\rho$, only $O(d)$ is taken into account because the permutation group and the translation group don't act on $\rho$. The equivariance of $SL_d(\Z)$ and $O(d)$ is trivial as $(g\cdot\rho)\cdot g'^{-1}=g\cdot(\rho\cdot g'^{-1})$ with $g\in\Euc(d)$ and $g'\in SL_d(\Z)$.\\
    Then, the permutation of $\Euc(d)$ and $SL_d(\Z)$ should be proven when the groups act on $x$. One can see that $O(d)$ doesn't act on the atomic positions $x$, consequently, we should only prove the permutation of the translation groups $E$ and $SL_d(\Z)$. We need to show that $\module{\module{ g'\cdot x}+(\rho\cdot g'^{-1})^{-1}\cdot v}=\module{ g'\cdot\module{ x+ \rho^{-1}\cdot v}}$, where the left member denotes the action of $SL_d(\Z)$ before $E$ while the right member denotes the action of $E$ before $SL_d(\Z)$. We get:
    \begin{align*}
        &\module{\module{ g'\cdot x}+(\rho\cdot g'^{-1})^{-1}\cdot v}\\
        =&\module{\module{ g'\cdot x}+g'\cdot\rho^{-1}\cdot v}\\
        =&\module{ g'\cdot x+g'\cdot\rho^{-1}\cdot v}
    \end{align*}
    And also:
    \begin{align*}
        &\module{ g'\cdot\module{ x+ \rho^{-1}\cdot v}}\\
        =&\module{ g'\cdot(x+ \rho^{-1}\cdot v)}\\
        =&\module{ g'\cdot x+ g'\cdot\rho^{-1}\cdot v}
    \end{align*}
    Consequently, the action of $SL_d(\Z)$ and $\Euc(\R)$ permutes.\\
    As $\Euc(d)$ and $SL_d(\Z)$ don't act on $z$ and have commutative action on $\rho$ and $x$, the two groups permute when they act on $\mathcal{M}^F_n$. Also, the equivariance of the permutation group is trivial as it acts on $x$ and $z$ by reordering the atomic positions $x_i$ and the chemical features $z_i$ while the other groups of $G$ act on the atomic positions without interacting with their order.
\end{proof}

\subsubsection*{Properties of the graph}
\begin{lemma}
    Consider $g\in\Euc(d)$ acting on $e_{ij\tau}$ by $g\cdot e_{ij\tau}=\module{x_i+\rho^{-1}\cdot v}-\module{x_j+\rho^{-1}\cdot v}+\tau$, then there exists a unique $\tau'$ in $\Z^d$ such that
    $$e_{ij\tau}=g\cdot e_{ij\tau'}.$$
\end{lemma}

\begin{proof}
    \begin{align*}
        e_{ij\tau}&=x_i-x_j+\tau\\
        &=\module{x_i-x_j}+\tau', \text{ with } \tau'\in\Z^d\\
        &=\module{x_i+\rho^{-1}\cdot v-x_j-\rho^{-1}\cdot v}+\tau'\\
        &=\module{\module{x_i+\rho^{-1}\cdot v}-\module{x_j+\rho^{-1}\cdot v}}+\tau'\\
        &=\module{x_i+\rho^{-1}\cdot v}-\module{x_j+\rho^{-1}\cdot v}+\tau'=g\cdot e_{ij\tau'}.
    \end{align*}
\end{proof}

\begin{lemma}
    Consider $g'\in SL_d(\Z)$ acting on $v_{ij\tau}$ by $g'\cdot v_{ij\tau}=\rho g'^{-1}(\module{g'\cdot x_i}-\module{g'\cdot x_j}+\tau)$, then there exists a unique $\tau'$ in $\Z^d$ such that
    $$v_{ij\tau}=g'\cdot v_{ij\tau'}.$$
\end{lemma}

\begin{proof}
    \begin{align*}
        v_{ij\tau}&=\rho\cdot(x_i-x_j+\tau)\\
        &=\rho\cdot(\module{x_i-x_j}+\tau'), \text{ with } \tau'\in\Z^d\\
        &=\rho\cdot(\module{\module{g'^{-1}\cdot g'\cdot x_i}-\module{g'^{-1}\cdot g'\cdot x_j}}+\tau')\\
        &=\rho\cdot(\module{g'^{-1}\cdot (\module{g'^{-1}\cdot x_i}-\module{g'^{-1}\cdot x_j})}+\tau')\\
        &=\rho\cdot(g'^{-1}\cdot (\module{g'\cdot x_i}-\module{g'^\cdot x_j})+\tau')\\
        &=\rho\cdot g'^{-1}\cdot(\module{g'\cdot x_i}-\module{g'\cdot x_j}+\tau')=g'\cdot v_{ij\tau'}
    \end{align*}
    because $\exists!\tau'\in\Z^d,(g\cdot x+\tau)=g\cdot (x+\tau')$ when $g\in SL_d(\Z)$ and $\tau\in\Z^d$
\end{proof}

\begin{definition}
    As $\exists!\tau'\in\Z^d,v_{ij\tau}=g'\cdot v_{ij\tau'}$ with $g\in\Euc(d)$, we can define the action of $\Euc(d)$ on $\tau$ as $g\cdot\tau=\tau'$. Moreover, as $\exists!\tau''\in\Z^d,v_{ij\tau}=g'\cdot v_{ij\tau''}$ with $g'\in SL_d(\Z)$, we can define the action of $SL_d(\Z)$ on $\tau$ as $g'\cdot\tau=\tau''$.
\end{definition}

As a result, we can see that $e_{ij\tau}$ is invariant to $\Euc(d)$ and $v_{ij\tau}$ is invariant to $SL_d(\Z)$ by definition. We can also extend the definition of the action of $SL_d(\Z)$ on $e_{ij\tau}$ as $g'\cdot v_{ij\tau}=\rho g'^{-1}(g'\cdot e_{ij\tau})\iff \rho e_{ij\tau}=\rho g'^{-1}(g'\cdot e_{ij\tau})\iff g'\cdot e_{ij\tau}=g'e_{ij\tau}$

\begin{lemma}
$||v_\gamma||$ and $\alpha_{\gamma\gamma'}$ are invariants to $G$.
\end{lemma}

\begin{proof}$\:$

\begin{itemize}
    \item with $g'\in SL_d(\Z)$:
    \begin{align*}
        &g'\cdot \alpha_{\gamma\gamma'}\\
        =&\atantwo(||(g'\cdot v_\gamma)\land (g'\cdot v_{\gamma'})||,(g'\cdot v_\gamma)\cdot (g'\cdot v_{\gamma'}))\\
        =&\atantwo(||v_\gamma\land v_{\gamma'}||,v_\gamma\cdot v_{\gamma'})\\
        =&\alpha_{\gamma\gamma'}
    \end{align*}
    \item with $g\in \Euc(d)$:\\
    \begin{align*}
        &g\cdot \alpha_{\gamma\gamma'}\\
        =&\atantwo(||(g\cdot v_\gamma)\land (g\cdot v_{\gamma'})||,(g\cdot v_\gamma)\cdot (g\cdot v_{\gamma'}))\\
        =&\atantwo(||g\cdot (v_\gamma\land v_{\gamma'})||,v_\gamma\cdot v_{\gamma'})\\
        =&\atantwo(||v_\gamma\land v_{\gamma'}||,v_\gamma\cdot v_{\gamma'})\\
        =&\alpha_{\gamma\gamma'}
    \end{align*}
\end{itemize}
\end{proof}

\section{Equivariant GNN for Materials}
\subsection{Equivariant Message Passing Neural Network}

\begin{lemma}\label{lemmamat}
    $$| g\cdot \rho e_\gamma \rangle \langle g\cdot\rho e_{\gamma'} | =g| \rho e_\gamma \rangle \langle \rho e_{\gamma'} |g^{-1} \text{ with } g\in\Euc(d)$$
\end{lemma}
\begin{proof}
    \begin{align*}
    | g\cdot \rho e_\gamma \rangle \langle g\cdot \rho e_{\gamma'} |
    =& g\rho e_\gamma e_{\gamma'}^\intercal\rho^\intercal g^\intercal\\
    =& g(\rho e_\gamma)(\rho e_{\gamma'})^\intercal g^\intercal\\
    =& g|\rho e_\gamma \rangle \langle \rho e_{\gamma'} | g^{-1}
    \end{align*}
   as  $g^\intercal=g^{-1}$ when  $g\in\Euc(d)$.
\end{proof}

Proposition 4 is true if $\lambda_{(g\cdot\gamma) (g\cdot\gamma')}=g \lambda_{\gamma\gamma'}g^{-1}$ when $g\in\Euc(d)$ and $\lambda_{(g\cdot\gamma) (g\cdot\gamma')}=\lambda_{\gamma\gamma'}$ when $g\in SL_d(\Z)$
\begin{proof}\label{proof_equivariance}
    First, we can observe that the actions on $z_i$ are invariant. Indeed, the definition of the messages $m_{ij\tau}$ is invariante as the geometrical information $||v_{ij\tau}||$ is invariante to the actions of $G$.

    As the actions of $\Euc(d)$ and $SL_d(\Z)$ are commutative, we can prove the equivarance of these two groups separately.
    
    We prove the equivariance of $\Euc(d)$ on $\rho$. It consists in showing:
    \begin{multline*}
    g\cdot\left(\exp(\sum_{(\gamma,\gamma') \in \Gamma_*} \ph_\theta^*(m_\gamma,m_{\gamma'},\alpha_{\gamma\gamma'}) \lambda_{\gamma\gamma'} )\rho\right)\\
    =\left(\exp(\sum_{(\gamma,\gamma') \in \Gamma_*} \ph_\theta^*(m_\gamma,m_{\gamma'},\alpha_{\gamma\gamma'}) \lambda_{(g\cdot\gamma)(g\cdot\gamma')} )g\cdot\rho\right),
    \end{multline*}
    with $\Gamma_*$ a set of edge pairs. It is an immediate application of Lemma \ref{lemmamat}.
    
    \begin{align*}
        &\left(\exp(\sum_{(\gamma,\gamma') \in \Gamma_*} \ph_\theta^*(m_\gamma,m_{\gamma'},\alpha_{\gamma\gamma'}) \lambda_{(g\cdot\gamma)(g\cdot\gamma')} )g\cdot\rho\right)\\
        =&\left(\exp(\sum_{(\gamma,\gamma') \in \Gamma_*} \ph_\theta^*(m_\gamma,m_{\gamma'},\alpha_{\gamma\gamma'}) g\lambda_{\gamma\gamma'}g^{-1} )g\cdot\rho\right)\\
        =&\left(\exp(g(\sum_{(\gamma,\gamma') \in \Gamma_*} \ph_\theta^*(m_\gamma,m_{\gamma'},\alpha_{\gamma\gamma'}) \lambda_{\gamma\gamma'})g^{-1} )g\cdot\rho\right)\\
        =&\left(g\exp(\sum_{(\gamma,\gamma') \in \Gamma_*} \ph_\theta^*(m_\gamma,m_{\gamma'},\alpha_{\gamma\gamma'}) \lambda_{\gamma\gamma'}) g^{-1}g\rho\right)\\
        =&g\cdot\left(\exp(\sum_{(\gamma,\gamma') \in \Gamma_*} \ph_\theta^*(m_\gamma,m_{\gamma'},\alpha_{\gamma\gamma'}) \lambda_{\gamma\gamma'} )\rho\right)\\
    \end{align*}
    
    Consequently, we can say that our proposed graph neural network is equivariant with the euclidian group $\Euc(d)$.
    
    Finally the equivariance on the unit cell $\rho$ is trivial as $v_{ij\tau}$ is invariant to $SL_d(\Z)$, consequently:
    \begin{align*}
    &\exp\left(\sum_{(\gamma,\gamma') \in \Gamma_*} \ph_\theta^*(m_\gamma,m_{\gamma'},\alpha_{\gamma\gamma'}) \lambda_{(g\cdot\gamma)(g\cdot\gamma')} \right)\rho\cdot g\\
    =&\left(\exp(\sum_{(\gamma,\gamma') \in \Gamma_*} \ph_\theta^*(m_\gamma,m_{\gamma'},\alpha_{\gamma\gamma'}) \lambda_{\gamma\gamma'} )\rho\right)\cdot g,
    \end{align*}
    Proving the equivariance and proposition 4.
\end{proof}

\subsection{EPGNN for Crystal Lattice Deformation}

\begin{equation}\label{message_formation}
    m^{l+1}_\gamma=\varphi^m_\theta(h^l_\src{\gamma}, h^l_\tgt{\gamma}, ||v^l_\gamma||)
\end{equation}
\begin{equation}\label{message_aggregation}
    h^{l+1}_i=\textbf{GRU}(h^l_i, \sum_{\gamma\in\Gamma(i)} m^{l+1}_{\gamma})
\end{equation}
\begin{equation}\label{lattice_deformation_1}
    w^{l+1}_{\gamma} = \varphi_\theta^{\rho^{(1)}}(h^{l+1}_\src{\gamma},h^{l+1}_\tgt{\gamma}, ||v^l_\gamma||)
\end{equation}
\begin{equation}\label{lattice_deformation_2}
    \begin{split}
    w^{l+1}_{\gamma \gamma'} = \varphi_\theta^{\rho^{(2)}}(&h^{l+1}_\src{\gamma},h^{l+1}_\tgt{\gamma},h^{l+1}_\tgt{\gamma'},\\
    &||v^l_\gamma||, ||v^l_{\gamma'}||, \angle^l(\gamma, \gamma'))
    \end{split}
\end{equation}
\begin{equation}\label{lattice_deformation}
    \rho^{l+1} = \left(I_3+k(\sum_{\gamma \in \Gamma_1} w^{l+1}_{\gamma} \lambda_\gamma+\sum_{(\gamma, \gamma') \in \Gamma_2} w^{l+1}_{\gamma, \gamma'} \lambda_{\gamma,\gamma'})\right) \cdot \rho^l
\end{equation}

%{\bf Remark.} 
In equation \ref{lattice_deformation}, we used a first order approximation of the exponential function on matrix.

\subsubsection*{Messages and weights.} The distance is encoded with a radial basis function $e_\text{RBF}:\R\to\R^D$ as $\tilde{e}_{\text{RBF},k}(d)=e^{-\frac{1}{\delta}(d-k\delta)^2}$ with $k\in\Iintv{0,n-1}$. The message function $\varphi^m_\theta:\R^F\times\R^F\times\R^+\to\R^F$ is defined as
\begin{equation}
    \varphi^m_\theta(h_i,h_j,d_{ij})=W'\silu{W[h_i||h_j||e_{\text{RBF}}(d_{ij})])}
\end{equation}
As $W\in\R^{(2F+D)\times F}$, $W'\in\R^{F\times F}$, with $\silu{x}=x\cdot\sigma(x)$ and $\sigma$ be the sigmoid function. In addition, let $||$ denotes the concatenation function. Then, the weighting of the triplets is defined by $\varphi^\rho_\theta$ as
\begin{equation}
    \varphi^{\rho^{(1)}}_\theta(h_i,h_j,d_{ij}) = {W^{(1)}}'\text{silu}(W^{(1)}[h_i||h_j||e_{\text{RBF}}(d_{ij})))
\end{equation}
\begin{equation}
    \begin{split}
        &\varphi^{\rho^{(2)}}_\theta(h_i,h_j,h_k,d_{ij},d_{ik},\alpha) = {W^{(2)}}'\text{silu}({W^{(2)}}[h_i||h_j||h_k\\
        &||e_{\text{RBF}}(d_{ij})||e_{\text{RBF}}(d_{ik})||\cos(\alpha)||\sin(\alpha)))
    \end{split}
\end{equation}
As ${W^{(1)}}\in\R^{(2F+D)\times F}$, ${W^{(2)}}\in\R^{(3F+2D+2)\times F}$ and ${W^{(1)}}',{W^{(2)}}'\in\R^F$.

\subsubsection*{Loss functions}
 The loss functions are as follows:

\begin{align*}
    \mathcal{L}^\text{Param}_\text{mae}=&\frac{1}{N}\sum^N_{k=1}\sum^6_{i=1}abs(p'_{ki}-p_{ki})\label{loss_mae_params}\\
    \mathcal{L}^\text{Param}_\text{mse}=&\frac{1}{N}\sum^N_{k=1}\sum^6_{i=1}(p'_{ki}-p_{ki})^2\\
    \mathcal{L}^\rho_\text{mae}=&\frac{1}{N}\sum^N_{k=1}\sum_{i,j}abs(F(\rho'_k)_{ij}-F(\rho_k)_{ij})\\
    \mathcal{L}^\rho_\text{mse}=&\frac{1}{N}\sum^N_{k=1}\sum_{i,j}(F(\rho'_k)_{ij}-F(\rho_k)_{ij})^2\\
    \mathcal{L}^\rho_\text{Riemann}=&\frac{1}{N}\sum^N_{k=1}trace(F(\rho'_k)F(\rho_k))
\end{align*}
with $F(A)$ being the metric tensor of $A$ as $F:\R^{3\times 3}\to\Sigma(3):A\mapsto A^\intercal A$.

\section{Experiments}

Up to 1715 experiments have been made on NVIDIA Tesla V100 GPU with. All experiments are done in less than 2 hours.

We used $\sigma=0.3$ for the experiments on loss function (in table 1) and $\sigma=0.1$ on the force field experiments (in table 2). The training is done with an Adam optimizer with a learning rate of $1e-4$, $3e-4$ and $1e-5$ during 32768 steps and a gradient clipping of $1.0$. The graph is built as a KNN graph with the 8 closest atoms. Our PEGNN model is composed of 6 layers of MPNN without lattice deformation and 4 layers with deformation with a feature space of size 128. The FFN baseline is composed of five layers with a hidden space of the same size as the feature space.

\subsubsection*{Grid search}

An Adam optimizer has been used during all experiments

\begin{tabular}{c|c}
    parameters & range \\
    \hline
    learning rate & $\{1e-4,3e-5,1e-5\}$\\
    batch size & 256\\
    total epoch & 32768\\
    knn graph & 8\\
    features space & 128\\
    weights scale & $\{\text{no limit}, 0.01\}$\\
    gradient clipping & 1.0\\
\end{tabular}

Weights scale denote a limit imposed over $w^{l+1}_{\gamma}$ and $w^{l+1}_{\gamma \gamma'}$ with a scaled sigmoid function.

\subsubsection*{DFT calculation}

Setting of the DFT calculation with VASP;

\begin{lstlisting}[caption=INCAR file]
SYSTEM = BFO_3D_no_magnetic

ISTART  = 0
IBRION  = 2
ISIF    = 6
NSW = 100
ENCUT    = 500
PREC     = Accurate
EDIFF    = 1e-7

LREAL = .FALSE.
ISMEAR   = 0
SIGMA    = 0.10
POTIM    = 0.020
LCHARGE  = .FALSE.
LWAVE    = .FALSE.

GGA = PE
\end{lstlisting}

\begin{lstlisting}[caption=KPOINTS file]
Automatic mesh
0              
Monkhorst-pack         
7   7   7        
0.  0.  0.  
\end{lstlisting}

\paragraph{Qualitative Analysis.} 

\begin{figure}[t]
\begin{subfigure}[b]{0.45\linewidth}
  \centering
  \textbf{samples of the structure with the lower ratio (succeeded samples)}
  \includegraphics[width=\columnwidth]{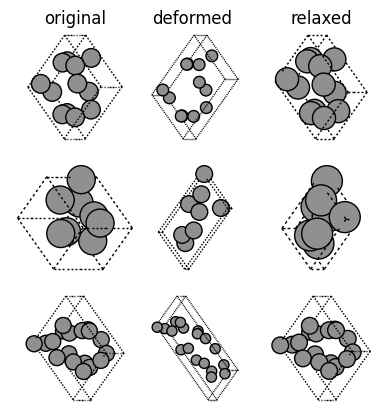}
  \caption{}
\end{subfigure} \hfill
\begin{subfigure}[b]{0.45\linewidth}
  \centering
  \textbf{samples of the structure with the highest ratio (failed samples)}
  \includegraphics[width=\columnwidth]{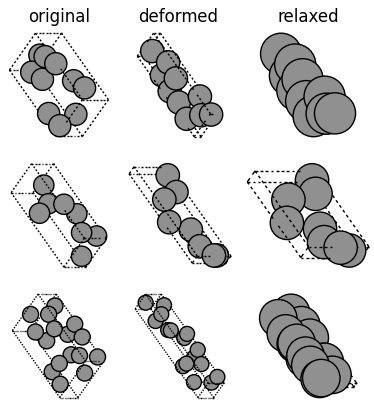}
  \caption{}
\end{subfigure}
  \caption{Relaxation applied by the PEGNN. The first column is the original structure, the second column is the deformed structure and the last column is the structure relaxed by the graph neural network. Each row corresponds to one sample.}
  \label{fig:exp}
\end{figure}

In figure \ref{fig:exp} (a), an important part of the deformed structure has been almost perfectly relaxed and reached a ratio below 0.1. As we can see, even more, important deformation of the unit cell can be well relaxed and the relaxed structure is much closer to the original structure than the deformed structures. On the other hand, some of the structures are not well relaxed as we can see in figure \ref{fig:exp} (b). However, these structures tend to have particular shapes. We can observe that a lot of failed relaxation starts from very elongated structure or has a unit cell with small sharp angles. In these cases, our model tends to shrink the structures. Such behaviours may be caused by the scarcity of short distance interactions, making these cases difficult to handle. In addition, a structure with a smaller relative distance may be more unstable since small variations of the geometry of the structure can have a greater impact on small crystals than large ones. One of the reasons why the crystals tend to elongate them-self is the periodicity of the graph. Indeed, if one or two dimensions are smaller that the other dimensions, an edge formed between a single atom and itself, a chain reaction can happened because these type of edge only acts on the lattice of the crystals. 

\begin{figure}[t]
  \centering
  \includegraphics[width=\columnwidth]{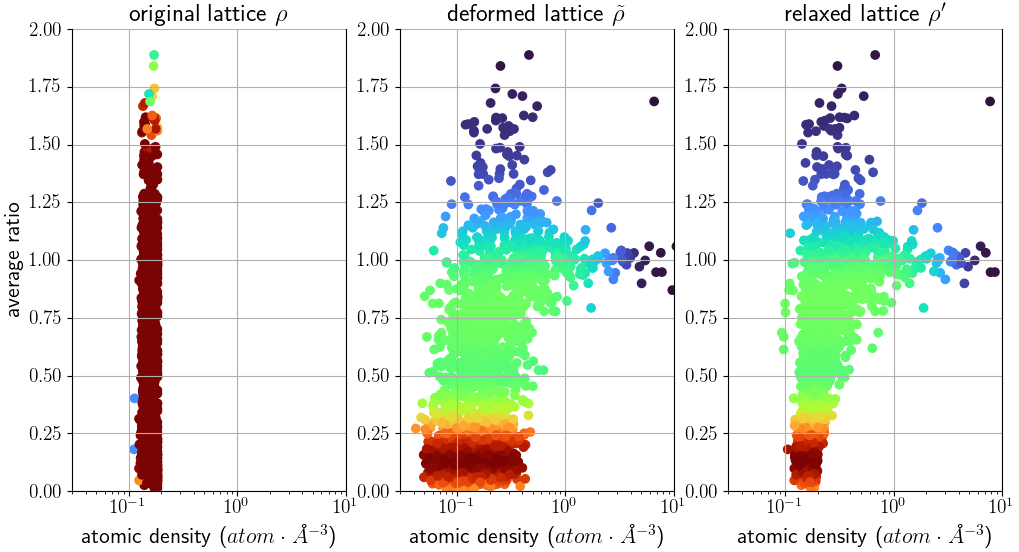}
  \caption{This figure represents the density of crystal samples on a heat-map according to the atomic density and the distance ratio from a given sample. An average ratio below one denote an improvement of the lattice parameters by the GNN. The first heat-map corresponds to the original lattice of the sample, the second to the deformed lattice and the third to the relaxed lattice.}
  \label{fig:density}
\end{figure}

To go further in our analysis, we have studied the relationship between the atomic density and the distance ratio of the samples as depicted in figure \ref{fig:density}. We can see that the atomic densities of the original lattices are all included in a narrower space than the deformed lattices. Moreover, the crystals with a high atomic density are not effectively relaxed by the graph neural network. We can deduce that our proposed GNN is not able to handle abnormally high density. As we can see, the sample density is high when the ratio is close to zeros. As a result, the case of failed relaxations seems to happen when the deformed crystal is too unrealistic. As the real dynamic simulation of the crystals, some deformed crystals may have reached a point where a realist simulation will not converge.

\end{document}